\PassOptionsToPackage{unicode}{hyperref}
\PassOptionsToPackage{hyphens}{url}
\PassOptionsToPackage{dvipsnames,svgnames,x11names}{xcolor}
\documentclass[
  onecolumn]{article}
\usepackage{xcolor}
\usepackage{amsmath,amssymb}
\usepackage{iftex}
\ifPDFTeX
  \usepackage[T1]{fontenc}
  \usepackage[utf8]{inputenc}
  \usepackage{textcomp} 
\else 
  \usepackage{unicode-math} 
  \defaultfontfeatures{Scale=MatchLowercase}
  \defaultfontfeatures[\rmfamily]{Ligatures=TeX,Scale=1}
\fi
\usepackage{lmodern}
\ifPDFTeX\else
\fi
\IfFileExists{upquote.sty}{\usepackage{upquote}}{}
\IfFileExists{microtype.sty}{
  \usepackage[]{microtype}
  \UseMicrotypeSet[protrusion]{basicmath} 
}{}
\makeatletter
\@ifundefined{KOMAClassName}{
  \IfFileExists{parskip.sty}{%
    \usepackage{parskip}
  }{
    \setlength{\parindent}{0pt}
    \setlength{\parskip}{6pt plus 2pt minus 1pt}}
}{
  \KOMAoptions{parskip=half}}
\makeatother
\makeatletter
\ifx\paragraph\undefined\else
  \let\oldparagraph\paragraph
  \renewcommand{\paragraph}{
    \@ifstar
      \xxxParagraphStar
      \xxxParagraphNoStar
  }
  \newcommand{\xxxParagraphStar}[1]{\oldparagraph*{#1}\mbox{}}
  \newcommand{\xxxParagraphNoStar}[1]{\oldparagraph{#1}\mbox{}}
\fi
\ifx\subparagraph\undefined\else
  \let\oldsubparagraph\subparagraph
  \renewcommand{\subparagraph}{
    \@ifstar
      \xxxSubParagraphStar
      \xxxSubParagraphNoStar
  }
  \newcommand{\xxxSubParagraphStar}[1]{\oldsubparagraph*{#1}\mbox{}}
  \newcommand{\xxxSubParagraphNoStar}[1]{\oldsubparagraph{#1}\mbox{}}
\fi
\makeatother

\usepackage{longtable,booktabs,array}
\usepackage{calc} 
\usepackage{etoolbox}
\makeatletter
\patchcmd\longtable{\par}{\if@noskipsec\mbox{}\fi\par}{}{}
\makeatother
\IfFileExists{footnotehyper.sty}{\usepackage{footnotehyper}}{\usepackage{footnote}}
\makesavenoteenv{longtable}
\usepackage{graphicx}
\makeatletter
\newsavebox\pandoc@box
\newcommand*\pandocbounded[1]{
  \sbox\pandoc@box{#1}%
  \Gscale@div\@tempa{\textheight}{\dimexpr\ht\pandoc@box+\dp\pandoc@box\relax}%
  \Gscale@div\@tempb{\linewidth}{\wd\pandoc@box}%
  \ifdim\@tempb\p@<\@tempa\p@\let\@tempa\@tempb\fi
  \ifdim\@tempa\p@<\p@\scalebox{\@tempa}{\usebox\pandoc@box}%
  \else\usebox{\pandoc@box}%
  \fi%
}
\def\fps@figure{htbp}
\makeatother

\NewDocumentCommand\citeproctext{}{}

\makeatletter
 \let\@cite@ofmt\@firstofone
 \def\@biblabel#1{}
 \def\@cite#1#2{{#1\if@tempswa , #2\fi}}
\makeatother
\newlength{\cslhangindent}
\newlength{\csllabelwidth}
\newenvironment{CSLReferences}[2] 
 {\begin{list}{}{%
  \setlength{\itemindent}{0pt}
  \setlength{\leftmargin}{0pt}
  \setlength{\parsep}{0pt}
  \ifodd #1
   \setlength{\leftmargin}{\cslhangindent}
   \setlength{\itemindent}{-1\cslhangindent}
  \fi
  \setlength{\itemsep}{#2\baselineskip}}}
 {\end{list}}
\usepackage{calc}

\usepackage{booktabs}
\usepackage{caption}
\usepackage{etoolbox}
\AtBeginEnvironment{longtable}{\footnotesize}
\makeatletter
\AtEndPreamble{%
  \let\c@definition\c@theorem \let\c@example\c@theorem
  \let\c@lemma\c@theorem      \let\c@proposition\c@theorem
  \let\c@corollary\c@theorem  \let\c@refremark\c@theorem}
\makeatother
\usepackage{arxiv}
\usepackage{orcidlink}
\usepackage{amsmath}
\makeatletter
\@ifpackageloaded{caption}{}{\usepackage{caption}}
\AtBeginDocument{%
\ifdefined\contentsname
  \renewcommand*\contentsname{Table of contents}
\else
  \newcommand\contentsname{Table of contents}
\fi
\ifdefined\listfigurename
  \renewcommand*\listfigurename{List of Figures}
\else
  \newcommand\listfigurename{List of Figures}
\fi
\ifdefined\listtablename
  \renewcommand*\listtablename{List of Tables}
\else
  \newcommand\listtablename{List of Tables}
\fi
\ifdefined\figurename
  \renewcommand*\figurename{Figure}
\else
  \newcommand\figurename{Figure}
\fi
\ifdefined\tablename
  \renewcommand*\tablename{Table}
\else
  \newcommand\tablename{Table}
\fi
}
\@ifpackageloaded{float}{}{\usepackage{float}}
\floatstyle{ruled}
\@ifundefined{c@chapter}{\newfloat{codelisting}{h}{lop}}{\newfloat{codelisting}{h}{lop}[chapter]}
\floatname{codelisting}{Listing}

\makeatother
\makeatletter
\@ifpackageloaded{caption}{}{\usepackage{caption}}
\@ifpackageloaded{subcaption}{}{\usepackage{subcaption}}
\makeatother
\usepackage{bookmark}
\IfFileExists{xurl.sty}{\usepackage{xurl}}{} 
\makeatletter
\@ifundefined{xmpquote}{\newcommand{\xmpquote}[1]{#1}}{}
\makeatother
\hypersetup{
  pdftitle={Structured Priors},
  pdfauthor={Thomas Bartz-Beielstein},
  pdfkeywords={\xmpquote{prior-data fitted
networks}, \xmpquote{TabPFN}, \xmpquote{in-context
learning}, \xmpquote{mixture priors}, \xmpquote{convolution of
priors}, \xmpquote{kernel mean embedding}, \xmpquote{Bayesian model
averaging}, \xmpquote{macl2l}},
  colorlinks=true,
  linkcolor={blue},
  filecolor={Maroon},
  citecolor={Blue},
  urlcolor={Blue},
  pdfcreator={LaTeX via pandoc}}

\renewcommand{\today}{2026-11-02}
\newcommand{\runninghead}{A Preprint }
\renewcommand{\runninghead}{Working paper }
\title{Structured Priors}
\usepackage{etoolbox}
\makeatletter
\providecommand{\subtitle}[1]{
  \apptocmd{\@title}{\par {\large #1 \par}}{}{}
}
\makeatother
\subtitle{Towards an algebra of component priors for prior-data fitted
networks}
\author{\textbf{Thomas
Bartz-Beielstein}~\orcidlink{0000-0002-5938-5158}\\\href{mailto:bartzbeielstein@gmail.com}{bartzbeielstein@gmail.com}}
\date{2026-11-02}
\begin{document}
\maketitle
\begin{abstract}
Prior-data fitted networks are typically trained using synthetic data
generated by a single, manually constructed generator. In cases where
such a generator combines multiple components, e.g., in TabPFN, Mitra,
or TabICLv2, this is achieved through a mixture using manually specified
weights, the impact of which on the network only becomes apparent after
pre-training. Instead of such a hand-built generator, this work proposes
a structured prior: a small set of named component priors, linked by
operators whose algebraic properties are formally proven and whose
influence on the trained network has been validated. Within a
measure-theoretic framework, the operators of mixture, convolution,
amplitude scaling, local transform, and projection are defined. Their
governing laws are proven. These operators are implemented in the
\texttt{macl2l} software package. Experiments demonstrate that the
implemented generators reproduce the identities within the resolution of
a permutation test. Furthermore, it is shown that trained networks obey
the distributive law; that a mixed prior composed of small components
performs as well as or better than any individual component in seven out
of eight open regression benchmarks; and that on two of these tables a
structured prior reaches, with a ninth of the data budget, an error that
a generic random-function prior does not reach even with the full
budget. A theorem on model averaging demonstrates that networks trained
on individual components---when averaged according to their prequential
likelihoods---constitute the Bayes predictor for the mixture, provided
that the component networks are themselves the Bayes predictors for
their respective priors. However, the pre-registered test of this
averaged predictor with trained component networks failed, because the
networks identify the generating component too rarely. The paper
proposes a research program focused on the algebra of structured priors
for prior-data fitted networks.
\end{abstract}
{\bfseries \emph Keywords}
\def\sep{\textbullet\ }
prior-data fitted networks \sep TabPFN \sep in-context
learning \sep mixture priors \sep convolution of priors \sep kernel mean
embedding \sep Bayesian model averaging \sep 
macl2l

\section{Introduction}\label{sec-intro}

Prior-data fitted networks (PFNs) are transformers trained on synthetic
datasets sampled from a prior over data-generating processes (Müller et
al., 2022). This approach was made practical for tabular data through
TabPFN (Hollmann et al., 2023; Hollmann et al., 2025). Models such as
TabICL, TabForestPFN, Mitra, and TabICLv2 incorporate tree-based and
other components into the generator (Breejen et al., 2024; Qu et al.,
2025, 2026; Zhang et al., 2025). In each case, however, these components
are combined via a categorical mixture or an ad hoc composition using
manually specified weights. Furthermore, the outcome of any modification
only becomes apparent after a pre-training run.

\begin{figure}[H]

\centering{

\includegraphics[width=1\linewidth,height=\textheight,keepaspectratio]{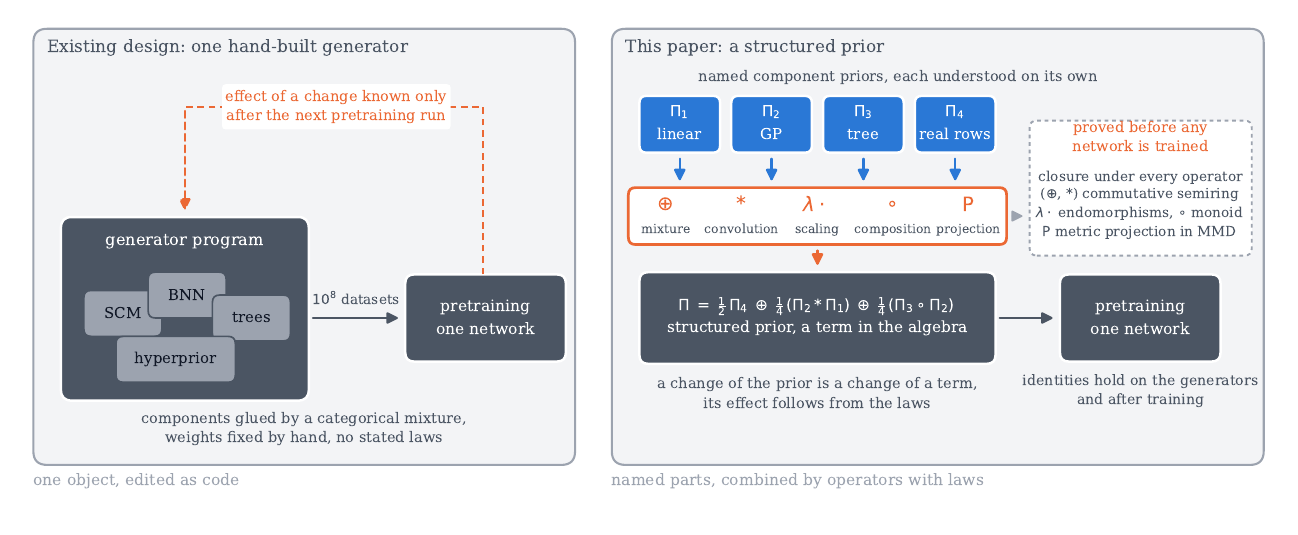}

}

\caption{\label{fig-structured-prior}One hand-built generator (left)
against a structured prior (right). On the left the components are glued
inside one program by a categorical mixture with weights fixed by hand.
The effect of a change is known only after a pretraining run. On the
right named component priors are combined by operators whose laws are
proved before any network is trained, so a structured prior is a term in
an algebra and a change of the prior is a change of a term.}

\end{figure}

The \emph{structured prior} proposed in this paper
(Figure~\ref{fig-structured-prior}) instead combines named component
priors using operators (see Section~\ref{sec-operators}) whose laws are
proved before any network is trained. In simple terms, a prior can be
thought of as a recipe for generating synthetic tables. This work treats
such recipes like numbers---they can be added, scaled, composed, and
compared. Three of its experiments were pre-registered: N1 tests whether
the implemented generators reproduce the laws of the algebra, N2 tests
the construction of one network per component, and N4 tests whether the
pruning by projection depends on the dataset kernel.

\section{Research questions and answers}\label{sec-rq}

This work addresses five research questions (RQ1--RQ5) concerning the
properties of priors, the operators used to combine them, and their
effects on the training of PFNs:

\subsection{RQ1: What constitutes a prior from which a program can draw,
and which types of priors can be constructed in this
manner?}\label{rq1-what-constitutes-a-prior-from-which-a-program-can-draw-and-which-types-of-priors-can-be-constructed-in-this-manner}

The paper defines a prior as the distribution of the function that a
generator produces and shows that these realisable priors form a cone
and that the pretraining objective of a PFN is minimised by the
posterior predictive distribution of its prior. Every object of the
theory can therefore be sampled, and the target of training is known.

\subsection{RQ2: Which operations combine priors such that the result is
again a prior---from which a program can sample---and what rules do
these operations
follow?}\label{rq2-which-operations-combine-priors-such-that-the-result-is-again-a-priorfrom-which-a-program-can-sampleand-what-rules-do-these-operations-follow}

It is shown that mixture and convolution turn the cone into a
commutative semiring. Amplitude scaling acts via endomorphisms of the
semiring, while local transforms form a monoid of additive maps that
distribute over mixtures. Projection onto a basis corresponds to the
metric projection onto the convex hull of the embedded basis with
respect to the Maximum Mean Discrepancy. A structured prior is thus a
term within an algebra, and modifying the prior corresponds to changing
a term.

\subsection{RQ3: Can multiple small networks---one for each
component---replace a single network trained on the
mixture?}\label{rq3-can-multiple-small-networksone-for-each-componentreplace-a-single-network-trained-on-the-mixture}

The question has a positive theoretical answer and a negative empirical
one. A model-averaging theorem shows that component networks, averaged
by their prequential likelihoods, are exactly the Bayes predictor of the
mixture when they are the Bayes predictors of their priors, and bounds
the cumulative excess risk of the averaged predictor otherwise by the
weighted average of the component risks. A corollary extends this
construction to include convolution, scaling, and transform applied to
the components, as well as the projection used to derive the weights. In
experiment N2, however, the trained component networks identify the
generating component too infrequently for this construction to be
practical.

\subsection{RQ4: Do the laws hold for the operators as implemented, and
do they remain valid after
training?}\label{rq4-do-the-laws-hold-for-the-operators-as-implemented-and-do-they-remain-valid-after-training}

A positive answer can be given here. Each operator is implemented as a
class or function within the Python software package \texttt{macl2l}
(Bartz-Beielstein, 2026). The identities hold for the implemented
generators, within the resolution of a permutation test. The
distributive law survives training.

\subsection{RQ5: Does a structured prior pay off compared to a
hand-constructed
one?}\label{rq5-does-a-structured-prior-pay-off-compared-to-a-hand-constructed-one}

For open regression tables, the answer is yes. A combination of real
windows and small ``teacher'' priors performs as well as or better than
any of its individual components on 7 out of 8 tables. On two of these
tables, a structured prior achieves a level of error using only
one-ninth of the data budget---a result that a generic random-function
prior fails to reach even with the full budget. The projection prunes a
basis of teachers at no measurable cost on the same two tables, but it
does not predict what adding a component gains, and the teachers it
admits depend on the dataset kernel.

\section{Operators and their laws}\label{sec-operators}

\begin{figure}[H]

\centering{

\includegraphics[width=1\linewidth,height=\textheight,keepaspectratio]{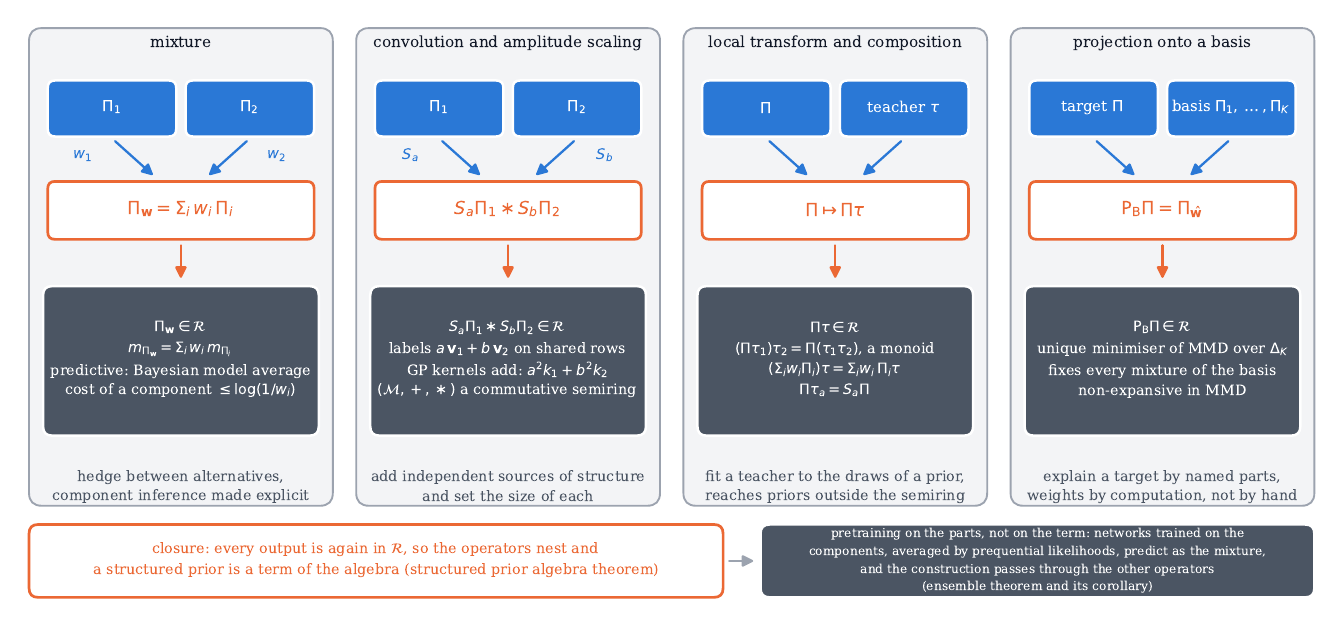}

}

\caption{\label{fig-operators-overview}The four operators of the paper,
one column each. Blue boxes are the inputs, the orange box is the
operator, and the dark box is its output together with the laws proved
for it; the grey line names the purpose. The mixture weights its
components. The convolution and the amplitude scaling add independent
draws and form, with the mixture, a commutative semiring. The local
transform fits a teacher to the draws of a prior, and transforms
compose. The projection returns the mixture of a basis that is closest
to a target in the maximum mean discrepancy. The strip at the bottom
holds the two payoffs shared by all operators: closure, collected in the
main theorem, and the averaging construction, which trains networks on
the parts rather than on the term and passes through the other
operators.}

\end{figure}

A prior designer who builds a prior from named parts wants to know,
before any network is trained, what the combination is, which laws it
obeys, and what it costs. The paper proves this for four operators. The
mixture picks one component at random and hedges between alternatives.
The convolution with amplitude scaling adds independent draws from
several priors and sets the size of each source of structure. A local
transform fits a data-driven teacher to the draws of a prior. Transforms
compose, so a teacher can be fitted to the output of another teacher.
The projection onto a basis explains a target prior by the closest
mixture of named priors. Two payoffs hold for all four. Every output is
again a realisable prior, so the operators nest and a structured prior
is a term of an algebra, which the main theorem of the paper collects.
Every operator comes with an exact law for the dataset law and the
predictive of its result, and the averaging theorem turns the law of the
mixture into a construction at inference time, which a corollary carries
through the other operators. Figure~\ref{fig-operators-overview} shows
the four operators with their inputs, outputs, laws, and purpose, and
Table~\ref{tbl-operators} lists the laws.

In Table~\ref{tbl-operators}, \(\Pi\) denotes a prior, the law of the
function that a generator produces, \(\mathcal{R}\) the set of
realisable priors, \(\mathcal{M}\) the cone of realisable measures,
\(\mathbf{w}\) a weight vector in the simplex \(\Delta_K\) over \(K\)
components, \(S_a\) the scaling by \(a\), \(\tau\) a local transform,
\(\mathsf{P}_{\mathsf{B}}\) the projection onto a basis \(\mathsf{B}\),
and \(m_\Pi\) the dataset density of \(\Pi\); Table~\ref{tbl-symbols}
lists every symbol with its type.

\begin{longtable}[]{@{}
  >{\raggedright\arraybackslash}p{(\linewidth - 8\tabcolsep) * \real{0.1455}}
  >{\raggedright\arraybackslash}p{(\linewidth - 8\tabcolsep) * \real{0.2182}}
  >{\raggedright\arraybackslash}p{(\linewidth - 8\tabcolsep) * \real{0.1455}}
  >{\raggedright\arraybackslash}p{(\linewidth - 8\tabcolsep) * \real{0.3091}}
  >{\raggedright\arraybackslash}p{(\linewidth - 8\tabcolsep) * \real{0.1818}}@{}}
\caption{The four operators of the paper with their maps, closure
results, laws, and purpose}\label{tbl-operators}\tabularnewline
\toprule\noalign{}
\begin{minipage}[b]{\linewidth}\raggedright
Operator
\end{minipage} & \begin{minipage}[b]{\linewidth}\raggedright
Map
\end{minipage} & \begin{minipage}[b]{\linewidth}\raggedright
Result
\end{minipage} & \begin{minipage}[b]{\linewidth}\raggedright
Law proved
\end{minipage} & \begin{minipage}[b]{\linewidth}\raggedright
Purpose
\end{minipage} \\
\midrule\noalign{}
\endfirsthead
\toprule\noalign{}
\begin{minipage}[b]{\linewidth}\raggedright
Operator
\end{minipage} & \begin{minipage}[b]{\linewidth}\raggedright
Map
\end{minipage} & \begin{minipage}[b]{\linewidth}\raggedright
Result
\end{minipage} & \begin{minipage}[b]{\linewidth}\raggedright
Law proved
\end{minipage} & \begin{minipage}[b]{\linewidth}\raggedright
Purpose
\end{minipage} \\
\midrule\noalign{}
\endhead
\bottomrule\noalign{}
\endlastfoot
mixture &
\((\Pi_1, \dots, \Pi_K, \mathbf{w}) \mapsto \Pi_{\mathbf{w}} = \sum_i w_i \Pi_i\)
& \(\Pi_{\mathbf{w}} \in \mathcal{R}\) & dataset density
\(\sum_i w_i m_{\Pi_i}\); predictive the Bayesian model average; cost at
most \(\log(1/w_i)\) per component; addition of the semiring & hedge
between alternatives; component inference made explicit \\
convolution and amplitude scaling &
\((\Pi_1, \Pi_2) \mapsto \Pi_1 \ast \Pi_2\); \(\Pi \mapsto S_a \Pi\);
\(\circledast_{\mathbf{a}}\) & in \(\mathcal{R}\) & commutative semiring
\((\mathcal{M}, +, \ast)\) with the \(S_a\) as endomorphisms; labels
\(a \mathbf{v}_1 + b \mathbf{v}_2\) on shared rows; kernels add for
Gaussian processes & add independent sources of structure; set the size
of each \\
local transform and composition & \(\Pi \mapsto \Pi\tau\) &
\(\Pi\tau \in \mathcal{R}\) & monoid under composition; additive and
positively homogeneous, so it distributes over mixtures;
\(\Pi\tau_a = S_a \Pi\) & fit a data-driven teacher to the draws of a
prior; reaches priors outside the semiring \\
projection onto a basis &
\(\Pi \mapsto \mathsf{P}_{\mathsf{B}} \Pi = \Pi_{\hat{\mathbf{w}}}\) &
\(\mathsf{P}_{\mathsf{B}} \Pi \in \mathcal{R}\) & unique minimiser of
the maximum mean discrepancy over \(\Delta_K\); fixes every mixture of
the basis; non-expansive & explain a target by named parts; weights
found by computation \\
\end{longtable}

\section{Contributions}\label{sec-contrib}

This work lays the foundation for a broader research program concerning
the algebra of component priors for PFNs. It contributes to the theory
by defining a prior as the distribution of the function generated by a
generator. Furthermore, it proves the laws of mixture, convolution,
amplitude scaling, local transform, and projection. These are
collectively termed the \emph{algebra of structured priors}. An
averaging theorem for networks trained on component priors is
introduced, along with its extension to the other aforementioned
operators. Each law is stated alongside its required assumptions. Every
operator is implemented as a class or function of the Python package
\texttt{macl2l}. Pre-registered experiments and training studies test
the laws on the implemented generators and after training, and measure
what the mixture, the data budget, and the projection gain. The failed
test of the averaged predictor is reported.

\section{Practical consequences}\label{sec-practice}

A practitioner who pretrains a network for a family of related tables
with \texttt{macl2l} first designs named components: real windows, which
are blocks of rows cut from the table at hand, and teacher recipes,
which fit a standard model to reference rows. The prior is then written
as a term that combines these components by the operators. A prior built
this way can be read, because it is a term over named parts, and the
effect of changing a part is known from the laws before any retraining.
The projection finds the weights of a basis of teachers and drops every
teacher with weight zero. On 8 of 8 comparisons, the arm trained with
the two admitted teachers was at most one seed spread above the arm with
all five; the seed spread is the spread between runs with different
random seeds.

The costs are as follows: every family of tables needs training from
scratch, whereas a pretrained model needs at most one fine-tuning run
per family. The studies ran one regression recipe on public tables, so
classification and other recipes are untested. One network per
component, averaged at use, is not yet practical, and the first remedy
to try is a density head, an output that gives a distribution over the
label.

The complete proofs, the full experimental protocol, and the list of
excluded candidates will be provided in a later version of this working
paper.

\section*{Notation}\label{notation}
\addcontentsline{toc}{section}{Notation}

\begin{longtable}[]{@{}
  >{\raggedright\arraybackslash}p{(\linewidth - 4\tabcolsep) * \real{0.2184}}
  >{\raggedright\arraybackslash}p{(\linewidth - 4\tabcolsep) * \real{0.5862}}
  >{\raggedright\arraybackslash}p{(\linewidth - 4\tabcolsep) * \real{0.1954}}@{}}
\caption{Symbols of the extended abstract, ordered Latin lower case,
Latin upper case, calligraphic and sans serif, Greek,
operators}\label{tbl-symbols}\tabularnewline
\toprule\noalign{}
\begin{minipage}[b]{\linewidth}\raggedright
Symbol
\end{minipage} & \begin{minipage}[b]{\linewidth}\raggedright
Meaning
\end{minipage} & \begin{minipage}[b]{\linewidth}\raggedright
Type
\end{minipage} \\
\midrule\noalign{}
\endfirsthead
\toprule\noalign{}
\begin{minipage}[b]{\linewidth}\raggedright
Symbol
\end{minipage} & \begin{minipage}[b]{\linewidth}\raggedright
Meaning
\end{minipage} & \begin{minipage}[b]{\linewidth}\raggedright
Type
\end{minipage} \\
\midrule\noalign{}
\endhead
\bottomrule\noalign{}
\endlastfoot
\(a, b\) & amplitude scales & scalar \\
\(\mathbf{a}\) & vector of amplitude scales of a weighted convolution &
vector \\
\(m_\Pi\) & dataset density of \(\Pi\) & density \\
\(\mathbf{v}_1, \mathbf{v}_2\) & vectors of clean labels on a row block
& vector \\
\(w_i, \mathbf{w}, \hat{\mathbf{w}}\) & mixture weights, weight vector
in \(\Delta_K\), projection weights & scalar, vector \\
\(K\) & number of component priors & integer \\
\(S_a\) & amplitude scaling of a measure by \(a\) & operator \\
\(\mathcal{M}\) & cone of realisable measures & set \\
\(\mathcal{R}\) & realisable priors & set \\
\(\mathsf{B}\) & basis of priors \((\Pi_1, \dots, \Pi_K)\) & tuple \\
\(\mathsf{P}_{\mathsf{B}}\) & projection onto the basis \(\mathsf{B}\) &
operator \\
\(\Delta_K\) & simplex of weight vectors & set \\
\(\Pi, \Pi_i\) & priors, in the algebra realisable priors & measure \\
\(\Pi_{\mathbf{w}}\) & mixture of \(\Pi_1, \dots, \Pi_K\) with weights
\(\mathbf{w}\) & measure \\
\(\Pi\tau\) & prior \(\Pi\) transformed by \(\tau\) & measure \\
\(\tau, \tau_a\) & local transform, amplitude transform & transform \\
\(+\) & addition of measures, the addition of the semiring &
operation \\
\(\ast\), \(\circledast_{\mathbf{a}}\) & convolution, weighted
convolution & operation \\
\end{longtable}

\section*{Use of AI tools}\label{use-of-ai-tools}
\addcontentsline{toc}{section}{Use of AI tools}

Transparency notice: parts of this paper were researched, drafted, and
checked with the support of artificial intelligence. This information is
provided voluntarily for complete transparency, as no legal requirement
mandates its disclosure.

\section*{References}\label{references}
\addcontentsline{toc}{section}{References}

\protect\phantomsection\label{refs}


\begin{CSLReferences}{1}{0}
\bibitem[\citeproctext]{ref-bart26bb}
Bartz-Beielstein, T. (2026). \emph{macl2l: A framework for efficient and
flexible in-context learning on {Apple} silicon} {[}Unpublished
manuscript{]}.

\bibitem[\citeproctext]{ref-bree24a}
Breejen, F. den, Bae, S., Cha, S., \& Yun, S.-Y. (2024).
\emph{Fine-tuned in-context learning transformers are excellent tabular
data classifiers}. \url{https://arxiv.org/abs/2405.13396}

\bibitem[\citeproctext]{ref-holl23a}
Hollmann, N., Müller, S., Eggensperger, K., \& Hutter, F. (2023).
{TabPFN}: A transformer that solves small tabular classification
problems in a second. \emph{The Eleventh International Conference on
Learning Representations ({ICLR})}.
\url{https://openreview.net/forum?id=cp5PvcI6w8_}

\bibitem[\citeproctext]{ref-holl25a}
Hollmann, N., Müller, S., Purucker, L., Krishnakumar, A., Körfer, M.,
Hoo, S. B., Schirrmeister, R. T., \& Hutter, F. (2025). Accurate
predictions on small data with a tabular foundation model.
\emph{Nature}, \emph{637}, 319--326.
\url{https://doi.org/10.1038/s41586-024-08328-6}

\bibitem[\citeproctext]{ref-muel21a}
Müller, S., Hollmann, N., Pineda Arango, S., Grabocka, J., \& Hutter, F.
(2022). Transformers can do {Bayesian} inference. \emph{International
Conference on Learning Representations ({ICLR})}.
\url{https://openreview.net/forum?id=KSugKcbNf9}

\bibitem[\citeproctext]{ref-qu25a}
Qu, J., Holzmüller, D., Varoquaux, G., \& Le Morvan, M. (2025).
{TabICL}: A tabular foundation model for in-context learning on large
data. \emph{Proceedings of the 42nd International Conference on Machine
Learning ({ICML})}. \url{https://arxiv.org/abs/2502.05564}

\bibitem[\citeproctext]{ref-qu26a}
Qu, J., Holzmüller, D., Varoquaux, G., \& Le Morvan, M. (2026).
\emph{{TabICLv2}: A better, faster, scalable, and open tabular
foundation model}. \url{https://arxiv.org/abs/2602.11139}

\bibitem[\citeproctext]{ref-zhan25c}
Zhang, X., Maddix, D. C., Yin, J., Erickson, N., Ansari, A. F., Han, B.,
Zhang, S., Akoglu, L., Faloutsos, C., Mahoney, M. W., Hu, C., Rangwala,
H., Karypis, G., \& Wang, B. (2025). Mitra: Mixed synthetic priors for
enhancing tabular foundation models. \emph{Advances in Neural
Information Processing Systems 38 ({NeurIPS})}.
\url{https://arxiv.org/abs/2510.21204}

\end{CSLReferences}
\end{document}